%% file: main.tex
\PassOptionsToPackage{table}{xcolor}
\documentclass[10pt,twocolumn,letterpaper]{article}
\usepackage{times}
\usepackage{cvpr}              % To produce the CAMERA-READY version

\definecolor{cvprblue}{rgb}{0.21,0.49,0.74}
\usepackage[pagebackref,breaklinks,colorlinks,allcolors=cvprblue]{hyperref}
\usepackage{hyperref}
\usepackage{url}
\usepackage{booktabs}
\usepackage{graphicx}
\usepackage{amsmath}
\usepackage{multirow}
\usepackage{wrapfig}
\usepackage{overpic}
\usepackage{float}
\usepackage{tablefootnote}
\usepackage{tabularx}

\def\paperID{11747} % *** Enter the Paper ID here
\def\confName{CVPR}
\def\confYear{2025}

\title{MMAR: Towards Lossless Multi-Modal Auto-Regressive Probabilistic Modeling}

\author{%
  \textbf{Jian Yang}\textsuperscript{1, 2,}~\footnotemark[1]~\textsuperscript{, }\footnotemark[4]%
  \quad%
  \textbf{Dacheng Yin}\textsuperscript{2,}\footnotemark[1]%
  \quad%
  \textbf{Yizhou Zhou}\textsuperscript{2,}\footnotemark[2]%
  \quad%
  \textbf{Fengyun Rao}\textsuperscript{2} \\
  \textbf{Wei Zhai}\textsuperscript{1}%
  \quad%
  \textbf{Yang Cao}\textsuperscript{1},%
  \quad%
  \textbf{Zheng-Jun Zha}\textsuperscript{1,}\footnotemark[3] \\
  \textsuperscript{1} MoE Key Laboratory of Brain-inspired Intelligent Perception and Cognition, \\
  \hspace*{1.5em}University of Science and Technology of China \\
  \textsuperscript{2} WeChat Vision, Tencent Inc. \\
  \small%
  \texttt{\{yangjian12138@mail., wzhai056@, forrest@, zhazj@\}ustc.edu.cn} \\
  \small%
  \texttt{\{dachengyin, fengyunrao\}@tencent.com}%
  \quad%
  \texttt{zyz0205@hotmail.com}%
}

\begin{document}

\maketitle
\renewcommand{\thefootnote}{\fnsymbol{footnote}}
\footnotetext[1]{Co-first Author}
\footnotetext[2]{Project Leader}
\footnotetext[3]{Corresponding Author}
\footnotetext[4]{Work done during an internship at WeChat Vision, Tencent Inc.}
\renewcommand{\thefootnote}{\arabic{footnote}} 

\input{sec/0_abstract}    
\input{sec/1_intro}
\input{sec/2_related_works}
\input{sec/3_method}
\input{sec/4_experiment}
\input{sec/5_conclusion}
\input{sec/ack}
{
    \small
    \bibliographystyle{ieeenat_fullname}
    \bibliography{main}
}

% WARNING: do not forget to delete the supplementary pages from your submission 
\input{sec/X_suppl}

\end{document}

%% file: sec/0_abstract.tex
\begin{abstract}
Recent advancements in multi-modal large language models have propelled the development of joint probabilistic models capable of both image understanding and generation. However, we have identified that recent methods suffer from loss of image information during understanding task, due to either image discretization or diffusion denoising steps. To address this issue, we propose a novel Multi-Modal Auto-Regressive (MMAR) probabilistic modeling framework. Unlike discretization line of method, MMAR takes in continuous-valued image tokens to avoid information loss in an efficient way. Differing from diffusion-based approaches, we disentangle the diffusion process from auto-regressive backbone model by employing a light-weight diffusion head on top each auto-regressed image patch embedding. In this way, when the model transits from image generation to understanding through text generation, the backbone model's hidden representation of the image is not limited to the last denoising step. To successfully train our method, we also propose a theoretically proven technique that addresses the numerical stability issue and a training strategy that balances the generation and understanding task goals. Extensive evaluations on 18 image understanding benchmarks show that MMAR significantly outperforms most of the existing joint multi-modal models, surpassing the method that employs pre-trained CLIP vision encoder. Meanwhile, MMAR is able to generate high quality images. We also show that our method is scalable with larger data and model size. Code will be available at \url{https://github.com/ydcUstc/MMAR}.
\end{abstract}

%% file: sec/1_intro.tex
\section{Introduction}
\label{sec:intro}

Over the past few years, extensive research in the field of multimodal intelligence has catalyzed the accelerated advancement of foundational models for both image understanding and image generation. Within the realm of image understanding, multimodal large language models (MLLM), exemplified by LLaVA~\citep{llava}, have exhibited human-like capabilities in open-domain image comprehension. In the domain of image generation, techniques rooted in generative probabilistic models, such as Denoising Diffusion Probabilistic Models (DDPM)~\citep{ddpm} and auto-regressive (AR) models~\citep{imagegpt}, have also garnered significant success. Essentially, these two lines of research correspond to modeling the conditional probability, i.e., $P(T|I)$ and $P(I|T)$, where $T$ and $I$ corresponds to text and image, respectively. It's evident that both types of conditional probabilistic models are subsets of a joint probabilistic model, $P(T,I)$. This brings us to an intriguing question: \textit{Could a joint probabilistic model serve as a natural unifying framework for both image understanding and image generation tasks?}

Given that the most advanced image understanding~\citep{internvl} and generation~\citep{sd3} models rely on the language priors $p(T)$ of pre-trained large language models (LLMs), the most straight-forward approach for joint image-text probabilistic modeling is to convert images into discrete tokens similar to text. This way, images are treated as a form of language and integrated into the auto-regressive modeling of text, as seen in models like MARS~\citep{mars}, LlamaGen~\citep{llamagen}, and Chameleon~\citep{chameleon}. This method leverages the powerful text modeling capabilities of various open-source pre-trained LLMs, along with their highly optimized training and inference frameworks. 
However, the code-book size strictly restricts the information capacity per token, leading to a loss of image details and reducing the model’s information capacity when the image token number is small. This limitation is quantitatively evident in the image understanding performance: most of the existing methods based on image token discretization~\citep{chameleon, show-o} fall short when compared to the LLaVA model, which utilizes continuous CLIP representations. 
Although some approaches can mitigate this bottleneck, such as increasing the number of tokens (4096 tokens per 512x512 image for EMU3~\citep{emu3}) or increasing the size of the VQ code-book, these approaches would substantially increase the training cost for both LLMs and VQ-VAE models. In contrast, without the restriction of code-book size, continuous image tokenizers can compress images much more efficiently. Recent research~\citep{dcae} even shows that it is possible to compress a 128x128 image patch into a single continuous image token without significant detail loss, resulting in only 16 tokens per 512x512 image. This indicates huge potential of using continuous image tokens instead of discrete ones.
\begin{figure}[t]
    \centering
    \includegraphics[width=\linewidth]{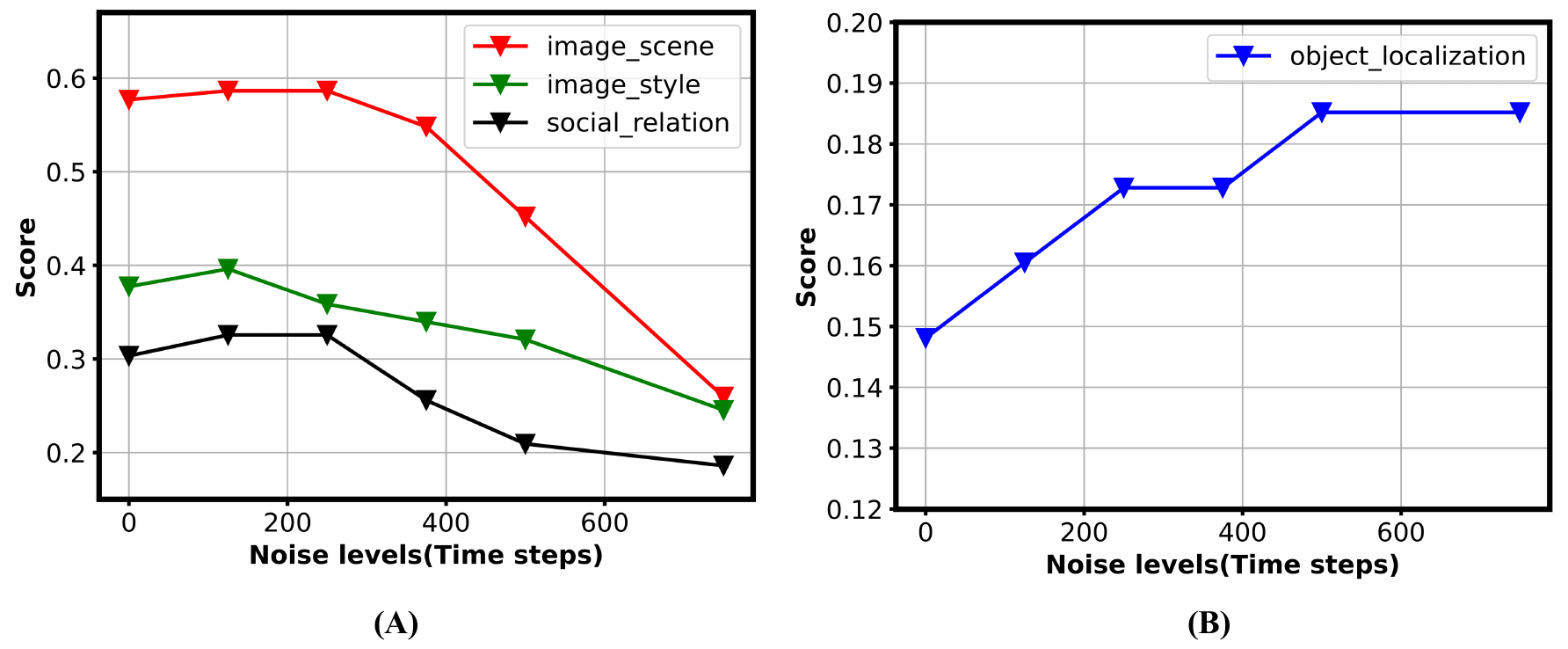}
    \vspace{-2em}
    \caption{The trend of MMBench tasks' performance of a transfusion-like model with respect to input image noise levels. (A): The best performance of most tasks are achieved at a non-zero noise level. (B): The object localization performance consistently gets higher as the noise level grows up.}
    \label{fig:transfusion_analysis1}
    \vspace{-1em}
\end{figure}

Recent efforts on leveraging continuous image tokens include transfusion~\citep{transfusion} and MonoFormer~\citep{monoformer}. They combine image diffusion transformers and text auto-regression within a unified transformer architecture, and demonstrates superior performance compared to image discretization approaches. 
However, when we reproduce and analyze this approach, we find it has difficulty utilizing complete image modeling capability for image understanding tasks when only the clean image is input to the model. As shown in Fig. \ref{fig:transfusion_analysis1}, the best performance of different image understanding tasks are achieved at different non-zero noise levels. Especially, the object localization task even prefers very noisy inputs. 
This counter-intuitive phenomenon reveals that, diffusion models encode different parts of image information into different noise levels. At low noise level, the model tends to solve an image enhancement task, while at high noise level, the model is only suitable for extracting rough image morphology and layout~\citep{ddpm}.
Intuitively, image understanding needs to take clean images as input to maintain complete visual information. However, due to the nature of diffusion modeling, this will only invoke the hidden representations that are related to the image enhancement task according to its low noise level, resulting in insignificant help for image understanding. Adding some noise to the input image would help invoke more semantic representations, but at a cost of image information loss. To ensure a full utilization of the image modeling capacity learned by image diffusion, it is necessary to input the images of all noise levels, but with tremendous computational overhead. This dilemma hinders transfusion-like models perform well for visual understanding tasks.

\begin{figure*}
    \centering
    \vspace{-1em}
    \includegraphics[width=0.99\linewidth]{imgs/intro_2.pdf}
    \caption{Comparison of different multi-modal probabilistic modeling paradigms. (A): Most AR-based methods adopt discrete image tokens produced by VQ-VAEs. The VQ operation bottlenecks per-tokens information capacity, resulting in information loss or huge token numbers. (B): Transfusion-like methods combining image DiT with text AR. $X^{(i)}$ denotes the full image at noise level $i$. For DiT part, the image information is modeled within the learned representation across various noise level, represented by $h^{(0)}, ..., h^{(t)}$, where the superscript indicates noise level. However, only $h^{(0)}$ is utilized in image to text tasks. (C): Our MMAR paradigm utilizes continuous image tokens $x_1, ..., x_n$, enjoying efficient and lossless token compression. Its auto-regressive nature guarantees that all the image hidden representations, denoted by $h_i, i\in[1, n]$, are simultaneously optimized for both $P(I)$ and $P(T|I)$, ensuring complete usage of image modeling capability for visual understanding tasks.}
    \label{fig:intro}
    \vspace{-0.5em}
\end{figure*}

In summary, \textbf{how to efficiently take in and utilize the complete information of the continuous image modality} is the major pain point of joint image-text probabilistic modeling. This is the challenge that our work is trying to address.
As shown in Fig. \ref{fig:intro}, the method proposed in this paper, MMAR, belongs to the image-text joint AR methodology, which ensures the complete use of image modeling capability by its auto-regressive nature. Different from other image-text joint AR methods, our method takes in continuous image tokens that can efficiently maintain the complete image information.
To model the continuous probability distribution within the auto-regressive paradigm, we refer to MAR~\citep{mar}, introducing a simple mlp diffusion sampler to sample continuous image tokens given the transformer backbone output as its conditional information. 
This technique can also be regarded as disentangling the diffusion process from the auto-regressive backbone in the transfusion-like methods~\citep{transfusion, monoformer}, which in turn focuses on learning complete visual representation that can guide the entire diffusion sampling process of image tokens.
Different from MAR~\citep{mar}, we leverage LLMs to learn much more diverse distribution of the large-scale image-text data. In practice, we find that low-precision training strategies for contemporary LLMs introduce non-negligible numerical error term in the diffusion sampler, which substantially impedes the convergence. We carefully analyze the numerical error term and find that a effective way to minimize it is to properly parameterize the diffusion model. 
Therefore, we theoretically derive an optimal diffusion parameterization to solve this issue. 
Additionally, visual understanding and generation prefer different experimental settings. To balance these two tasks in a unified model, we propose a two-stage training method, which firstly improves visual understanding ability with large, mid quality data, then boost visual generation ability with high quality data. 
With these innovations, the lossless multi-modal auto-regressive probabilistic modeling is finally achieved theoretically and practically.

Our contributions are the following three folds:
\begin{itemize}
    \item We identified why previous joint probabilistic models for both image generation and understanding suffer image information loss.
    \item We are the first to combine continuous image representation and discrete text representation into a unified auto-regressive probabilistic modeling framework. Our framework successfully overcomes information loss while keeping high efficiency, leading to significant performance gains on various image understanding tasks. 
    \item We proposed two training techniques crucial for training the model: one solves numerical error issue under low-precision training setting with theoretical guarantee, the other balances image generation and understanding tasks goals.
\end{itemize}

%% file: sec/2_related_works.tex
\section{Related Works}
\label{sec:related_work}
\subsection{Multi-modal Large Language Models}
Since LLMs demonstrated open-domain conversational capabilities, more and more research has been focused on how to introduce visual information into large language models to achieve open-domain visual understanding. Pioneering works such as BLIP-2~\citep{blip2}, MiniGPT4~\citep{minigpt4} and LLaVA~\citep{llava} use trainable connector modules such as qformer or mlp to align image representations to the input space of LLM, making open-domain visual question answering possible. Recently, thanks to innovations in network architecture~\citep{xcomposer,cambrian}, training strategy~\citep{internvl} and the support for dynamic resolution input~\citep{minicpm,llava_next}, the visual understanding performance of large multi-modal models have been greatly improved. These works focus on the alignment of image representations to text representations, only achieving $p(T|I)$ without incorporating the image's own distribution $p(I)$ into the modeling capabilities of the model. Different from these works, our work adds the modeling of $p(I)$ on the basis of these MLLMs to achieve joint image-text probabilistic modeling.

\subsection{Auto-Regressive Image Generative Models}
Text-to-image generative models aim at modeling the conditional probability distribution $p(I|c)$, enabling probabilistic sampling of images conditioned on textual or categorical inputs. Auto-regressive methods represent a dominant paradigm in this domain, typically requiring discrete representations for both input and output. 
For images, this necessitates encoding them into discrete codes using a VQVAE~\citep{vqgan, videopoet}. While recent works demonstrate that auto-regressive methods based on discrete image tokens can generate high-quality images~\citep{llamagen}, the discretization of image representation acts as an information bottleneck, limiting the modeling accuracy of the image distribution. Recent efforts have shown that auto-regressive probabilistic modeling can be achieved without relying on discrete representations~\citep{givt, mar}. For instance, MAR~\citep{mar} replaces traditional logits with diffusion heads, enabling probabilistic sampling of continuous representations within an auto-regressive framework.
This paper introduces continuous representation auto-regressive probability modeling to MLLMs, mitigating information loss caused by quantization and achieving theoretically lossless joint image-text probability modeling. In addition, we addressed the difficulty when training with large model and large data which is not presented in the previous works.

\subsection{Unified Visual Understanding and Generation Models}
Recently, a series of studies have focused on leveraging a single model to simultaneously address tasks of visual generation and understanding. Early works in this area adopted a modular approach, bridging pre-trained models for visual understanding and visual generation through intermediate representations to achieve combined image understanding and generation. 
Notable examples include EMU~\citep{emu, emu2} and SEED-X~\citep{seedx}. 
These works, however, are not considered probabilistic modeling because they aim at modeling the mean value of representations like CLIP or other intermediate representations rather than modeling the true image distribution. 
This limitation leads to the inadequate image space exploration, and thus hinders the attainment of high-quality generative and understanding performance.

Another line of research adheres to the paradigm of probabilistic modeling~\citep{chameleon, mars, show-o, vila-u, transfusion, monoformer}. These approaches can be categorized into three types based on whether the image representations are discrete and the modeling method of the image part :
(i) Discrete auto-regressive methods: Examples include Chameleon~\citep{chameleon}, MARS~\citep{mars}, VILA-U~\citep{vila-u}, and EMU-3~\cite{emu3}. These methods discretize image representations and then model images and text tokens using a unified auto-regressive transformer.
(ii) Discrete diffusion methods: An example is Show-o~\citep{show-o}. These methods discretize image tokens and model them with text tokens using a unified transformer through a discrete diffusion approach.
(iii) Continuous diffusion methods: Examples include Transfusion~\citep{transfusion} and MonoFormer~\citep{monoformer}. These methods do not discretize image representations but directly employ continuous diffusion transformers to model image tokens along with text tokens using a unified transformer.
Our approach differs from the aforementioned three types. It belongs to the continuous auto-regressive method category, which does not require discretizing image representations. Instead, it predicts the continuous distribution of image tokens using an auto-regressive approach and models them alongside text tokens within a unified transformer.

%% file: sec/3_method.tex
\section{Method}
\label{method}
\subsection{Auto-Regressive Modeling with Continuous and Discrete Representations}
Auto-regressive modeling is a commonly used probabilistic modeling method for sequence data. It can be formulated by ``predicting the next token" as follows:
\begin{equation}
    \log p_\theta(\mathbf{x}) = \sum_{i=1}^{n}\log p_\theta(x_{i}|x_{<i}),
\end{equation}
where $\theta$ and $\mathbf{x}$ represent model parameters and the sequence data, respectively. By maximizing the log likelihood of the data, $\mathbb{E}_{\mathbf{x}\sim\mathcal{D}}\log p_\theta(\mathbf{x})$, the model can be optimized to sample from the data distribution $\mathcal{D}$, achieving probabilistic modeling.

In the realm of natural language processing (NLP), the sequence $\mathbf{x}$ is solely made of discrete text tokens. As a result, most modern large language models (LLMs) parameterize $p_\theta(x_{i}|x_{<i})$ into a categorical distribution, which can be explicitly represented by the softmax activation on a set of logits predicted by a decoder-only transformer~\citep{gpt2, gpt3} $f_\theta(\cdot)$ followed by a linear LM head $H_\theta(\cdot)$:
\begin{equation}
    p_\theta(x_{i}|x_{<i}) = \mathrm{softmax}(H_\theta(f_\theta(x_{<i}))).
\end{equation}
In addition to text, our work also aims at modeling the probability of images, which are represented by continuous rather than discrete image tokens. Therefore, a protocol for parameterizing $p_\theta(x_{i}|x_{<i})$ of the continuous image tokens is required. Inspired by MAR~\citep{mar}, 
we train a diffusion model to achieve this. The diffusion model takes vector $z_i=f_\theta(x_{<i})$ as the conditional input, and samples $x_{i}$ by gradually denoising from a randomly sampled gaussian noise. To optimize the diffusion model for continuous image token sampling, a diffusion loss can be applied, which acts as the upper-bound of the negative log likelihood. A typical diffusion loss can be written as follows, which is seen in MAR~\citep{mar}:
\begin{align}
     L_i &= \mathbb{E}_{x_{i},\epsilon,t}[w_t \cdot ||\mathbf{\epsilon} - \mathbf{\epsilon}_\theta(\sqrt{\bar{\alpha}_t}x_i + \sqrt{1 - \bar{\alpha}_t}\mathbf{\epsilon}, t, z_i)||^2] \notag\\
     &\ge -\log p_\theta(x_{i}|x_{<i}) + C, \label{eq:n-pred}
\end{align}
where $w_t$ is the loss weight that balances the loss for different timesteps, and $\bar{\alpha}_t$ indicates the noise schedule of the forward diffusion process. In this way, minimizing the diffusion loss is equivalent to maximizing the log likelihood of image data conditioned on its preceding text guidance. 

The overall loss for joint image-text probabilistic modeling can be written as follows, which is an upper-bound of the negative log likelihood of the multi-modal data $\mathbf{x}$:
\begin{equation}
     L = \sum_{i\in I_{img}}L_i - \sum_{i\in I_{txt}}\log p_\theta(x_{i}|x_{<i}) \ge - \log p_\theta(\mathbf{x}) + C,
\end{equation}
where $I_{img}$ and $I_{txt}$ indicate the indices of image tokens and text tokens, respectively.
\subsection{Optimal Diffusion Model Parameterization for Low-Precision Training}
In the era of large language models, training with low precision data type, such as \texttt{bfloat16}, has become increasingly popular. However, the training and inference process of a diffusion model is relatively sensitive to numerical precision, we provide direct empirical demonstration in Appendix \ref{app:demo_numerical}. Moreover, in an auto-regressive framework, the image tokens are sampled sequentially, requiring even more precise sampling for each image token to reduce the overall error accumulation. Therefore, handling the numerical error in the diffusion process modeling should be emphasized when integrating diffusion loss into LLMs. 

\begin{figure*}
    \centering
    \vspace{-1em}
    \includegraphics[width=0.95\linewidth]{imgs/Method.pdf}
    \vspace{-0.5em}
    \caption{The overview of MMAR with two stage image expert training strategy.}
    \label{fig:Model_Architecture}
    \vspace{-1.7em}
\end{figure*}

From the example below, we can clearly illustrate the effect of the numerical error in diffusion models. In Eq. \ref{eq:n-pred}, the diffusion model is parameterized as $\mathbf{\epsilon}_\theta(\sqrt{\bar{\alpha}_t}x_i + \sqrt{1 - \bar{\alpha}_t}\mathbf{\epsilon}, t, z_i)$, known as ``$\epsilon$-prediction", predicting the noise $\epsilon$ that is added to the data.
Floating-point representation causes numerical error's amplitude proportional to a number's magnitude. This can be modeled by scaling the prediction by a factor of $(1 + \delta)$, where $\delta$ is the relative error. 
In this way, we can write DDIM~\citep{ddim} sampling with numerical error as follows:
\begin{align}
     \Tilde{x}^{(t-1)} =& \sqrt{\bar{\alpha}_{t-1}}\left(\frac{x^{(t)} - \sqrt{1-\bar{\alpha}_t}\epsilon_\theta(x^{(t)},t, z_i)(1+\delta)}{\sqrt{\bar{\alpha}_t}}\right) \notag\\
     &+ \sqrt{1-\bar{\alpha}_{t-1}}\epsilon_\theta(x^{(t)},t, z_i)(1+\delta). \label{eq:ddim-numerical-err}
\end{align}
Further separating the numerical error term from the ideal DDIM sampling process, we get:
\begin{align}
     \Tilde{x}^{(t-1)} =& (\sqrt{1-\bar{\alpha}_{t-1}} -\frac{\sqrt{\bar{\alpha}_{t-1}}}{\sqrt{\bar{\alpha}_{t}}}\sqrt{1-\bar{\alpha}_t})\epsilon_\theta(x^{(t)},t, z_i)\delta \notag\\
     &+ x^{(t-1)} ,
\end{align}
where the first term is the numerical error term, and the second term $x^{(t-1)}$ is the ideal DDIM sampling term. Let $\sigma_t$ be the standard deviation of $\epsilon_\theta(x^{(t)},t, z_i)$, then the standard deviation of numerical error term can be calculated as $|\sqrt{1-\bar{\alpha}_{t-1}} -\frac{\sqrt{\bar{\alpha}_{t-1}}}{\sqrt{\bar{\alpha}_{t}}}\sqrt{1-\bar{\alpha}_t}|\delta\sigma_t$. When the signal-to-noise ratio (SNR) is high, i.e. $\bar{\alpha}_t, \bar{\alpha}_{t-1} \to 1$, the numerical error has almost zero amplitude. However, when SNR is extremely low, i.e. $\bar{\alpha}_t = 0$, and $\bar{\alpha}_{t-1} > 0$, this term will explode to infinity, causing extreme numerical error.

Our goal is to minimize the numerical error term. To achieve this, we analyze most of the factors that can determine the numerical error in Appendix \ref{app:v-pred}, and find that an effective solution is to parameterize the diffusion model properly. By solving the numerical error minimization problem, we conclude that the v-prediction parameterization~\citep{vpred} is the desired optimal parameterization method. Note that v-prediction is initially proposed for the efficient distillation of diffusion models, rather than reducing the numerical error of diffusion models. To the best of our knowledge, our work is the first to derive  v-prediction parameterization from the first principle of minimizing the numerical error in diffusion models. For more details, see Appendix \ref{app:v-pred}. 
Under v-prediction parameterization, the model predicts a linear combination of data $x_i$ and noise $\epsilon$:
\begin{equation}
     v^{(t)}_i = \sqrt{\bar{\alpha}_t}\mathbf{\epsilon} - \sqrt{1 - \bar{\alpha}_t}x_i. \label{eq:v-pred}
\end{equation}
We therefore re-write Eq.\ref{eq:n-pred} into the v-prediction form, and set the loss weight $w_t$ to 1 for simplicity:
\begin{equation}
     L_i = \mathbb{E}_{x_{i},\epsilon,t}[||v_i^{(t)} - v_\theta(\sqrt{\bar{\alpha}_t}x_i + \sqrt{1 - \bar{\alpha}_t}\mathbf{\epsilon}, t, z_i)||^2]. \label{eq:v-pred-loss}
\end{equation}

\subsection{Implementation}
\subsubsection{Model Design}
Overall, our model follows the auto-regressive (AR) modeling paradigm for both image and text. For the text part, we leverage a pre-trained decoder-only LLM~\citep{qwen2} for the modeling of text probability $P(T)$. While for the image part, we pick the masked AR implementation~\citep{mar}, which has been demonstrated to perform much better than decoder-only transformer implementation, meanwhile maintaining the form of $p_{\theta}(x_i|x_{<i})$ modeling.
Our design goal is to embed the masked AR into the pre-trained LLM to realize a unified $f_{\theta}(\cdot)$ for multi-modal input and output. Generally, a masked AR implementation includes a bi-directional encoder, which encodes the known image tokens $x_{<i}$, and a bi-directional decoder, which decodes the masked image token $x_i$ given $x_{<i}$ and the mask position embeddings. Our approach is shown in Fig. \ref{fig:Model_Architecture}. Firstly, we introduce an EmbeddingViT module to serve as the role of the encoder. Secondly, we directly embed the decoder into the LLM. This is achieved by adding PLoRA-based~\citep{xcomposer} image expert into the LLM, and setting the attention of the image region to be bi-directional. We also set the same ROPE position ID for all image tokens to ensure that only the mask position embedding decides the image token position information. In this way, the decoder-only LLM behaves just like the masked AR decoder when the PLoRA is activated. Meanwhile, thanks to the pre-trained LLM, the decoding of image tokens can be conditioned on the preceding text. To calculate the image diffusion loss, a Diffusion MLP module is adopted. This module takes the embedding of the image part output by the LLM, denoted by $z_i=f_{\theta}(x_{<i})$, as a conditional input to predict $v_i^{(t)}$ according to Eq. \ref{eq:v-pred}, ultimately achieving the loss of Eq. \ref{eq:v-pred-loss}. For detailed module specifics, please refer to Appendix \ref{app:VQ_Transfusion}. 

During training, we simultaneously learn two tasks: image understanding and image generation. For image understanding task, the sequence is arranged as \texttt{[img][text]}, and we perform text auto-regressive training under the condition of all image tokens, achieving $P(T|I)$. For conditional image generation task, the sequence is arranged as \texttt{[text][img]}, and we perform image masked AR training under the condition of preceding text tokens, achieving $P(I|T)$. Specifically, the image tokens are randomly masked with a controlled mask ratio decided by the training strategy. For unconditional image generation task, the sequence is arranged as \texttt{[img]} only, and we perform image masked AR training without any preceding text conditions, achieving $P(I)$. This also allows the application of Classifier-Free Guidance (CFG) techniques. Combining $P(I|T)$, $P(T|I)$, $P(I)$, and $P(T)$, we achieve the modeling of $P(I, T)=P(I | T)P(T)=P(T|I)P(I)$, form both text to image and image to text direction.

\subsubsection{Two Stage Training}
Image understanding tasks are often applied in various complex scenarios. Therefore, the training requires highly diverse data, including but not limited to blurred images, small-sized images, images with greatly differing aspect ratios, etc. However, image generation tasks, on the other hand, place higher demands on data clarity, resolution, and even aesthetic criteria is one of the ways to evaluate generative performance.
To achieve a balance between image generation and understanding capabilities, our training process is carried out in two stages. The first stage, called Image Expert Pretraining stage, utilizes large-scale, mid-quality data (as illustrated in Fig. \ref{fig:Model_Architecture}) to enhance the diversity of the model's data distribution~\citep{minicpm}. By initially modeling this diverse data, we strengthen the model's understanding capabilities. In the second stage, called Image Expert Fine-tuning stage, we employ a smaller volume of high-quality data to further improve the image generation capacity and refine the model's comprehension of images.

\begin{figure}[t]
    \centering
    \vspace{0 em}
    \includegraphics[width=0.85\linewidth]{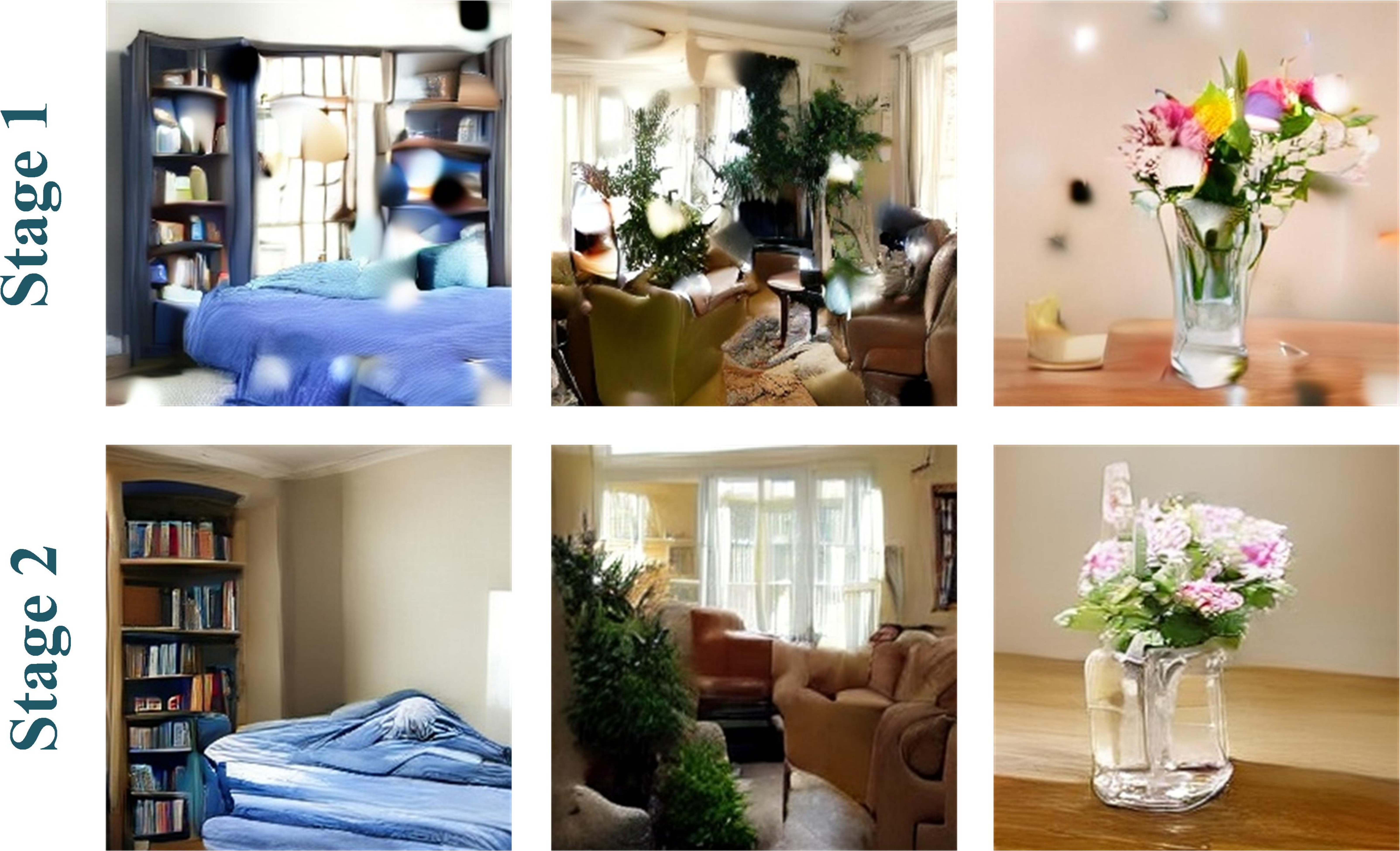}
    \vspace{-0.5em}
    \caption{Training stage 2 with (0,1] mask ratio range addresses the issue of generating holes of training stage 1.}
    \label{fig:hole}
    \vspace{-1.0 em}
\end{figure}

Surprisingly, we find that the images generated by the model at the end of the first training stage all have holes, as shown in Fig. ~\ref{fig:hole}. We control the step-by-step generation of images tokens and ultimately find that the appearance of holes always occurs within the last thirty percent of the generation sequence. And this is because the first stage uses a [0.7, 1.0] mask ratio, which in the context of fewer training epochs, results in the model still not adapting to the image completion scene task of the last thirty percent. To address this issue, we set the lower bound of the mask ratio to 0 in the second stage and adjust the variance of the mask ratio distribution to enable the model to learn image generation tasks in various scenarios as efficiently as possible. The probability density curve of the two mask ratios are shown on the right side of Fig. \ref{fig:Model_Architecture}. Ultimately, this achieves a rapid improvement in the model's generative capabilities within a limited number of training epochs.
\begin{table*}[t]
  \centering
  \vspace{-1em}
  \caption{Comparison on visual understanding benchmarks. MMAR surpasses other joint image-text probabilistic models even with a small resolution of 256x256, approaching the performance of traditional MLLMs like LLaVA, which employ pre-trained CLIP vision encoder.}
  \vspace{-0.5em}
  \fontsize{7.5pt}{7.5pt}\selectfont
  \begin{tabular}{cllll|ccccccc}
    \toprule
     \textbf{Type} & \textbf{Method} & \textbf{LLM} & \textbf{V-Token} & \textbf{Res.} & \textbf{MMB} & \textbf{MME$^P$} & \textbf{POPE} & \textbf{SEED} & \textbf{MM-Vet} & \textbf{AVE@18Und.} \\
    \midrule
    \multirow{1}{*}{Und. Only} & LLaVA-1.5~\citep{llava1.5} & Vicuna-1.5-7B & CLIP & 336 & 64.3 & \textbf{1510.7} & \textbf{85.9} & 58.6 & 31.1 & {\color{blue}47.08} \\
    \midrule
    \multirow{2}{*}{\shortstack{Und. \& Gen. \\Non-prob.}} & EMU-2~\citep{emu} & LLaMA-13B & CLIP & 448 & -- & -- & -- & 62.8 & \textbf{48.5} & --\\
    ~ & SEED-X~\citep{seedx} & LLaMA-13B & CLIP & dynamic & \textbf{75.4} & 1435.7 & 84.2 & -- & -- & --\\
    \midrule
    \multirow{7}{*}{\shortstack{Joint Prob. \\Models}} & Chameleon-7B~\citep{chameleon} & 7B from scratch & vq-vae & 512 & 13.32  & 125.8 & 30.86 & 34.61  & 7.34 & 18.34 \\
    ~ & Transfusion* & Qwen-2-0.5B & vae & 256 & 29.47  & 594.3 & 66.90 & 42.40 & 13.90 & 28.26 \\
    ~ & Show-o~\citep{show-o} & Phi-1.5B & CLIP & 336 & 42.44  & 1182.7 & 84.50 & 51.61 & 20.87 & 33.06 \\
    ~ & VILA-U~\citep{vila-u} & LLaMA-2-7B & vq-vae & 256 & -- & 1336.2 & 83.9 & 56.3 & 27.7 & -- \\
    ~ & EMU-3~\citep{emu3} & 8B from scratch & vq-vae & 512 & 58.5 & -- & {\color{blue}85.2} & {\color{blue}68.2} & {\color{blue}37.2} & -- \\
    % \midrule
    \rowcolor{gray!25}
    ~ & MMAR-0.5B & Qwen-2-0.5B & vae & 256 & 48.45  & 882.1 & 70.74  & 55.70  & 18.49  & 34.56 \\
    \rowcolor{gray!25}
    ~ & MMAR-7B & Qwen-2-7B & vae & 256 & {\color{blue}70.45}  & {\color{blue}1486.9} & 84.02  & \textbf{68.63}  & 30.64  & \textbf{48.25} \\
    \bottomrule
  \end{tabular}
  \vspace{-1.5em}
  \label{UndMetrics}
\end{table*}

%% file: sec/4_experiment.tex
\section{Experiment}
\label{experiment}
\subsection{Dataset}
We utilized the Capfusion-120M dataset~\citep{capsfusion} for the image expert pretraining stage. This dataset is publicly accessible and comprises an extensive collection of web-based image-text pairs, designed to optimize noisy captions. In an effort to further improve the quality of the content generated, we executed a random sampling of 20M data points from the Capfusion dataset during our image expert fine-tuning stage. This was supplemented with a high-quality mixed dataset that included CC12M~\citep{cc12m}, and LAION-aesthetics-12M\footnote{https://huggingface.co/datasets/dclure/laion-aesthetics-12m-umap}. We employ the open-source InternVL2-8B~\citep{internVL_8b} for recaptioning the CC12M and laion-aesthetics-12m datasets in English. Following LLaVA-v1.5~\citep{llava}, we use LLaVA-v1.5-mix-665K data for instruction tuning before each performance test for visual understanding.

\subsection{Implementation Details}
By default, we used the AdamW~\citep{adamw} optimizer with betas (0.9, 0.95) and weight decay proportional to the learning rate. In the first training stage, a learning rate of 5e-5 was used for 4 epochs (with a 0.5-epoch warm-up) for the 0.5B model, and 2e-5 for 3 epochs (with a 0.5-epoch warm-up) for the 7B model. The second stage used 5e-5 for 3 epochs (with a 0.5-epoch warm-up) for the 0.5B model, and 2e-5 for 3 epochs (with a 0.5-epoch warm-up) for the 7B model. Batch sizes were 2496 for the 0.5B model and 1152 for the 7B model in the first stage, and 768 for the 0.5B model and 480 for the 7B model in the second stage. The image tokenizer and original pre-trained LLM parameters were frozen in both stages, while other parameters were trainable. A consistent image resolution of 256x256 and the LDM-KL-16~\citep{LDM} tokenizer were used throughout.

\subsection{Comparison with Other Systems}

\paragraph{Visual Understanding. }In Table \ref{UndMetrics}, we employ VLMEvalKit~\citep{vlmevalkit} to perform extensive evaluations on prevalent visual understanding benchmarks, encompassing a total of 18 such assessments (average score denoted by ``AVE@18Und." in Table \ref{UndMetrics}, for details see Appendix \ref{app:details_understand}.) including MMB~\citep{mmbench}, MME~\citep{mme}, POPE~\citep{pope}, SEED~\citep{seedbench}, MM-Vet~\citep{mmvet}, among others. 
Our method outperforms other joint image-text probabilistic models by a large margin, including Chameleon-7B, Show-o, VILA-U and our re-implemented version of Transfusion (denoted by ``Transfusion*"). With merely 256 image token input, it achieves comparable or better performance compared to EMU3, which needs 4096 tokens per image. Even without using any pre-trained CLIP or diffusion models and with small resolution of 256 $\times$ 256, MMAR-7B presents comparable or better performance when compared to methods rely on pre-trained clip and diffusion models, including understanding-only methods like LLaVA-1.5, and non-probabilistic joint image understanding and generation models like EMU-2 and SEED-X. 

\paragraph{Visual Generation.} We showcase the zero-shot FID~\citep{FID} of the MMAR at the end of the second stage training, evaluated on the MSCOCO 30k dataset~\citep{MSCOCO}, in Table \ref{FIDMetrics}. It is worth noting that, even with only 3 epochs of training and using a less pure high-quality dataset due to the need to balance understanding performance, MMAR has been able to generate reasonable images, although it has not achieved very good generation effects. Furthermore, in order to verify that MMAR has the ability to generate high-quality images, we additionally train for 6 epochs on a pure high-quality image dataset as the third stage, where we replace Capfusion with JourneyDB~\citep{JourneyDB} based on the second stage dataset. In the third stage, we only use image diffusion loss to further explore the model's generative capabilities. We evaluate MMAR on FID@MJHQ-30K~\citep{mjhq} and GenEval~\citep{GenEval} benchmarks at the end of the third stage training, and show results in Table \ref{FIDMetrics_stg3}. It can be seen that after training with pure high-quality images, our model's performance is discernibly on par with existing methods.
Meanwhile, our image loss is still decreasing, which indicates that the image generation capability can still be improved. Notably, we also evaluated the visual understanding ability of the MMAR at the end of the third stage, and the overall average metric only decreased by 1\%. For generated example images and visual understanding performance, please refer to Appendix \ref{app:gen_sample} and \ref{app:details_understand}.

\begin{table}[h]
  \centering
  \vspace{-1em}
  \caption{Ablation study on MMAR}
  \vspace{-0.5em}
  \fontsize{7.0pt}{7.0pt}\selectfont
  \begin{tabular}{l|ccc|c}
    \toprule
     \textbf{Exp. Setting}  & \textbf{MMB} & \textbf{MME$^P$} & \textbf{AVE@18Und.}  & \textbf{FID-30K$\downarrow$} \\
    \midrule
    \rowcolor{gray!25}
    MMAR-0.5B & \textbf{48.45}  & \textbf{882.1}  & \textbf{34.56} & \textbf{36.6}\\
    \midrule
    ~ w/ $\epsilon$-pred. & 45.53  & 880.7 & 32.21 & 61.53 \\
    \midrule
    ~ show-O like (w/VQ) & 37.54 & 618.2 & 29.70 & 66.26\\
    ~ transfusion-like & 29.47  & 594.3  & 28.26 & 95.38 \\
    \bottomrule
  \end{tabular}
  \vspace{-1em}
  \label{ModelAblation}
\end{table}

\subsection{Ablation Study}
Table \ref{ModelAblation} demonstrates that switching to the more common $\epsilon$-prediction leads to a significant decrease in both visual understanding and generation quality, confirming the effectiveness of our optimal diffusion parameterization.
To evaluate the efficacy of various image-text joint modeling techniques, we devise two distinct versions based on the MMAR framework: one employing LDM-VQ-16 tokenizer~\citep{LDM} for discrete token modeling (refer to show-O), and the other utilizing Transformer for diffusion modeling (refer to Transfusion). Implementation details are provided in the Appendix \ref{app:VQ_Transfusion}. Our test results are presented in Table \ref{ModelAblation}. The transfusion-like version considerably underperforms the other two approaches in both understanding and generation aspects. This is attributed to the substantial loss of image information when jointly modeling image-text and operating with limited training epochs. In contrast, our full method consistently delivers superior results.

\begin{table}[t]
  \centering
  \vspace{-0.5em}
  \caption{Comparison on MSCOCO Dataset}
  \vspace{-0.5em}
  \fontsize{7pt}{7pt}\selectfont
  \begin{tabular}{ccp{0.6cm}p{0.65cm}c}
    \toprule
     \textbf{Type} & \textbf{Method} & \textbf{Params} & \textbf{Images} & \textbf{FID-30K$\downarrow$} \\
    \midrule
    \multirow{4}{*}{Gen. Only} & DALL-E~\citep{dalle} & 12B & 250M &  27.50 \\
    ~ & LDM~\citep{LDM} & 1.4B & 400M & 12.64  \\
    ~ & DALL-E2~\citep{dalle2} & 6.5B & 650M & 10.39 \\
    ~ & Imagegen~\citep{imagegen} & 3B & 5000M+ & 6.61 \\
    \midrule
    
    \multirow{3}{*}{ \shortstack{Und. and Gen. \\ w/ pre-trained Diff.}}  & CoDI~\citep{codi} & - & 400M & 11.26 \\
    ~ & SEED-X~\citep{seedx} & 17B & - & 12.68 \\
    ~ & DreamLLM~\citep{dreamllm} & 7B & - & 8.76 \\
    \midrule
    \multirow{5}{*}{ Joint Prob. Models } & Show-o~\citep{show-o}  & 1.3B & 35M & 9.24 \\
    ~ & Chanmeleon~\citep{chameleon}  & 7B & - & 29.6 \\
    ~ & Transfusion~\citep{transfusion} \tablefootnote{Both the Transfusion and Chanmeleon results are referenced from Table 3 in the paper `Transfusion: Predict the Next Token and Diffuse Images with One Multi-Modal Model.'} & 7B & - & 16.8 \\
    \rowcolor{gray!25}
    ~ & MMAR-0.5B & 0.5B & 145.2M &36.6 \\
    \rowcolor{gray!25}
    ~ &  MMAR-7B & 7B & 145.2M  &  22.9   \\
    \bottomrule
  \end{tabular}
  \vspace{-2em}
  \label{FIDMetrics}
\end{table}

\subsection{Analysis}
\begin{figure}[h!]
    \centering
    \vspace{0.5em}
    \begin{overpic}[width=1.0\linewidth]{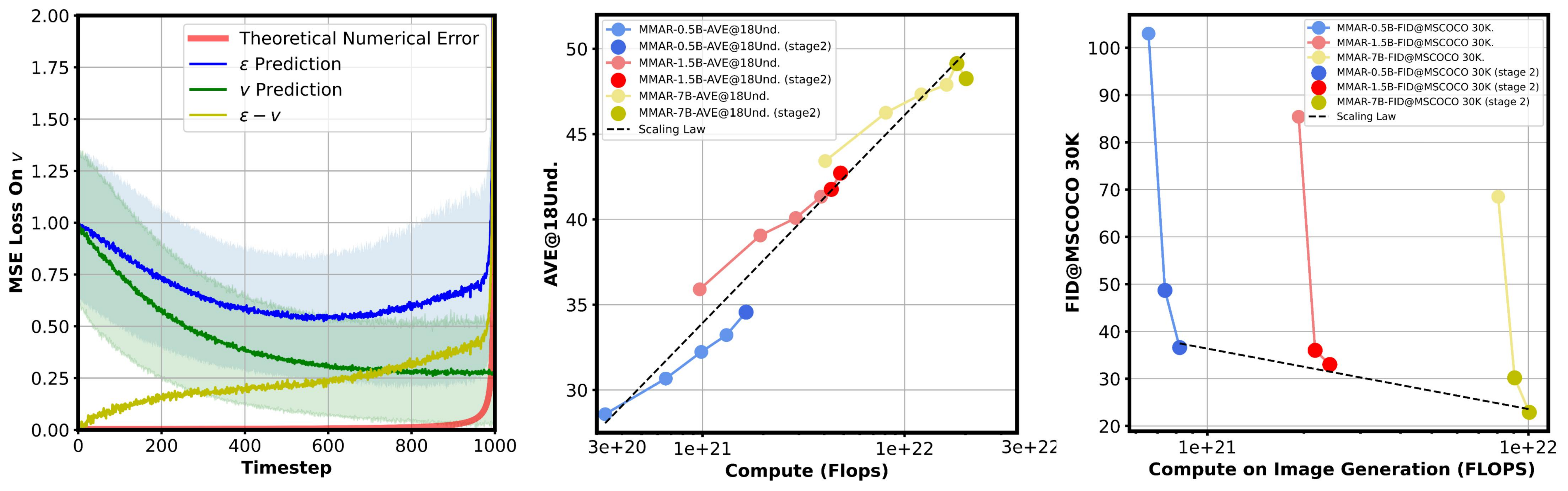} 
        \put(15,-4){\small(A)} 
        \put(49,-4){\small(B)} 
        \put(83,-4){\small(C)} 
    \end{overpic}
    % \vspace{0.2em}
    \caption{The impact of v-prediction and the scaling laws on the tasks of image understanding and generation.}
    \label{fig:analyze}
    % \vspace{-1.5em}
\end{figure}

\paragraph{Impact of v-prediction}
To delve deeper into the disparities between v-prediction and $\epsilon$-prediction for diffusion MLP, we independently collect the statistics of the MSE loss of $v^{(t)}$ values at various timesteps $t$ throughout the training process for both methods, as illustrated in Fig. \ref{fig:analyze} (A). 
Furthermore, to effectively discern the loss discrepancies between the two techniques, we plot the difference between v-prediction and the $\epsilon$-prediction curve, yielding the yellow curve. 
For reference, we plot the theoretical numerical error of the $\epsilon$-prediction model (details in Appendix \ref{app:theoretical_error}) as the red curve.
The graph reveals that the loss of $\epsilon$-prediction model is consistently higher than v-prediction model. Especially, when $t>900$, curve ``$\epsilon-v$" exhibits a significant spike towards infinity. This aligns with the behavior of the theoretical numerical error, confirming the non-negligible numerical error effect in low-precision training. The gap between yellow and red curve indicates that apart from its direct effect, numerical error also introduces optimization difficulty, hindering the loss convergence. 

\begin{table}[t]
  \centering
  \vspace{-0.5em}
  \caption{Comparison on MJHQ-30K Dataset and GenEval.}
  \vspace{-0.5em}
  \fontsize{7pt}{7pt}\selectfont
  \begin{tabular}{cccc}
    \toprule
     \textbf{Type} & \textbf{Method} & \textbf{FID$\downarrow$} & \textbf{GenEval overall$\uparrow$} \\
    \midrule
    \multirow{3}{*}{Gen. Only} & PixArt~\citep{pixart}  & 6.14 &  0.48 \\
    ~ & SDXL~\citep{sdxl}  & 9.55 & 0.55  \\
    ~ & SD3~\citep{sd3}  & - & 0.68 \\
    \midrule
    
    \multirow{2}{*}{ \shortstack{Und. and Gen. \\ w/ pre-trained Diff.}}  & CoDI~\citep{codi}  & - & 0.31 \\
    ~ & SEED-X~\citep{seedx}  & - & 0.55 \\
    \midrule
    \multirow{4}{*}{ Joint Prob. Models } & Show-o~\citep{show-o}  & 15.18 & 0.53 \\
    ~ & VILA-U~\citep{vila-u}   & 12.81  & - \\
    ~ & Transfusion~\citep{transfusion}   & - & 0.63 \\
    ~ & EMU3~\citep{emu3}   & - & 0.66 \\
    \rowcolor{gray!25}
    ~ &  MMAR-7B  & 15.6  &  0.51   \\
    \bottomrule
  \end{tabular}
  \vspace{-1em}
  \label{FIDMetrics_stg3}
\end{table}

\paragraph{Scaling Law}
We conduct scaling up experiments by varying model size (N) and training token count (D), approximating FLOPs as 6ND. Fig. \ref{fig:analyze} (B), (C) displays the performance on image understanding and generation benchmarks. The scaling laws are evident in both tasks, as shown by the dashed lines.

%% file: sec/5_conclusion.tex
\section{Discussion and Conclusion}
MMAR is the first to achieve the unified autoregressive probabilistic modeling for continuous image and discrete text representation. It addresses numerical instability and balances image comprehension and generation, thereby ensuring practicality and scalability. While masked autoregressive modeling serves as one of the design options for MMAR, our framework can be adaptable to future breakthroughs in continuous image autoregressive modeling.
MMAR uses continuous image tokens, boosting efficiency sixteenfold over image discretization, matching or exceeding EMU-3's image comprehension capabilities with fewer image tokens (4096 tokens/image to 256 tokens/image). This advantage grows with advanced continuous image tokenizers. And it would definitely benefit more when it comes to modeling of interleaved multi-image data and long videos.
MMAR is currently in early exploration, being experimented on relatively limited resolution, data scale, model size, and did not utilize more advanced continuous image tokenizers. Despite this, MMAR shows promise through its scalable performance and strong comprehension/generation abilities. Future works include unlocking these limitations and developing more advanced continuous image AR implementation other than MAR.

%% file: sec/ack.tex
\clearpage
\section*{Acknowledgments}
This work is supported by the National Natural Science Foundation of China (NSFC) under Grants 62225207, 62436008 and 62306295. Additionally, we thank Ziheng Zhou for his helpful discussion.

%% file: sec/X_suppl.tex
\clearpage
\setcounter{page}{1}
\maketitlesupplementary

\section{Appendix}
\label{sec:appendix}
\subsection{Additional Implementation Details}\label{app:VQ_Transfusion}

\begin{table}[h]
    \centering
    \vspace{-1em}
    \caption{Hyper-parameter settings for MMAR models}
    \vspace{-0.5em}
    \fontsize{8pt}{8pt}\selectfont
    \begin{tabular}{c|ccc}
    \toprule
     ~ & ~  & \textbf{MMAR-0.5B}  & \textbf{MMAR-7B}   \\
    \multirow{-2}{*}{\textbf{Module}}  & \multirow{-2}{*}{\textbf{Param.}}  & \textbf{settings} & \textbf{settings}  \\
    \midrule
     \cellcolor{gray!5}~  & \cellcolor{gray!10}$N_{diff}$   & \cellcolor{gray!10}8  & \cellcolor{gray!10}12  \\
     \cellcolor{gray!5}~ & $d_{mlp}$ & 1024 & 2048 \\
     \cellcolor{gray!5}\multirow{-3}{*}{\textbf{Diffusion MLP}} & \cellcolor{gray!10}$r_{mlp}$ & \cellcolor{gray!10}4 & \cellcolor{gray!10}4 \\
    \midrule
      \cellcolor{gray!5}~  & Res.  & 256  & 256 \\
     \cellcolor{gray!5}\multirow{-2}{*}{\textbf{Image Tonkenizer}} & \cellcolor{gray!10}$d$ &  \cellcolor{gray!10}16 &  \cellcolor{gray!10}16 \\
    \midrule
      \cellcolor{gray!5}  ~  & $N_{ViT}$  & 16  & 16  \\
      \cellcolor{gray!5}\multirow{-2}{*}{\textbf{EmbeddingViT}} & \cellcolor{gray!10}$d_{ViT}$  & \cellcolor{gray!10}1024  & \cellcolor{gray!10}1024 \\
    \midrule
      \cellcolor{gray!5}~ & $r$ & 512  &  1280 \\
      \cellcolor{gray!5}\multirow{-2}{*}{\textbf{LLM's PLoRA}} & \cellcolor{gray!10}$\alpha$ & \cellcolor{gray!10}128 & \cellcolor{gray!10}128 \\
    \bottomrule
    \end{tabular}
    \vspace{-2em}
    \label{tab:module_details}
\end{table}

\paragraph{Diffusion MLP.}

\begin{figure}[t]
    \centering
    \includegraphics[width=\linewidth]{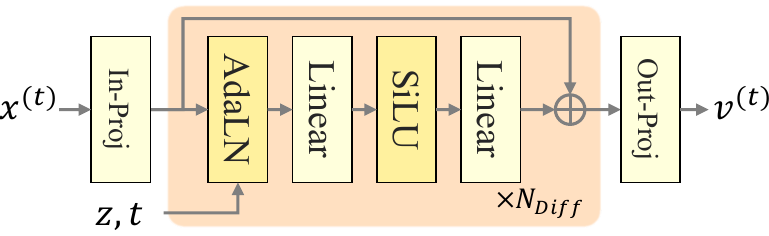}
    \caption{Details of Diffusion MLP architecture.}
    \label{fig:DiffMLP_details}
    \vspace{-2em}
\end{figure}

Inspired by MAR, we employ a simple MLP architecture to predict $v^{(t)}$, whose detailed architecture is shown in Fig. \ref{fig:DiffMLP_details}.
It consists of a main network and its input/output linear projection layers. The input projection converts the $d$-dimensional noisy latent $x^{(t)}$ into the mlp hidden dimension $d_{mlp}$, while the output projection converts the $d_{mlp}$-dimensional main network output back to $d$ dimensions.
The main network is a stack of $N_{diff}$ residual blocks, each comprising an AdaLN~\citep{AdaLN}, followed by a two-layer MLP activated by SiLU. The expand ratio of each MLP is denoted by $r_{mlp}$, which means its first linear layer projects from $d_{mlp}$ to the dimension of $d_{mlp} \times r_{mlp}$.
The condition vector $z$ is added to the diffusion time embedding and is incorporated through AdaLN. 
In practice, $r_{mlp}$ is set to 4. For MMAR-0.5B, $N_{diff}=8$ and $d_{mlp}=1024$, while for MMAR-7B, $N_{diff}=12$ and $d_{mlp}=2048$.

\paragraph{Image Tokenizer.} 
MMAR employs the publicly available KL-16 image tokenizer from LDM~\citep{LDM}. This tokenizer processes an 256 $\times$ 256 image into 256 image tokens, each with $d=16$ channels, and it is kept frozen during training. 

\paragraph{EmbeddingViT.} 
The EmbeddingViT module is implemented with a ViT\citep{vit} encoder with $N_{ViT}=16$ layers and $d_{ViT}=1024$ hidden state channels, processing image tokens into visual embeddings with stronger context awareness. We integrate a learnable position embedding for each image token in EmbeddingViT, corresponding to its 2D position on the image.

\paragraph{LLM.} 
We initialize our LLM using parameters from the open-source QWen2 series models~\citep{qwen2}. To preserve text modeling capabilities, we keep the LLM parameters fixed during training and add PLoRA~\citep{xcomposer} as the image expert, where only the image tokens in the input pass through the introduced LoRA~\citep{lora} adapters. The PLoRA is applied for each linear layer within the original LLM. Considering the time cost, our ablation study employs QWen2-0.5B-Instruct, with $r=512$ and $\alpha=128$ for PLoRA. Furthermore, we use QWen2-7B-Instruct to explore our method's scale-up capability with $r=1280$ and $\alpha=128$ for PLoRA. The corresponding MMAR models are denoted by MMAR-0.5B and MMAR-7B, respectively.

\paragraph{Projector.} 
We use two MLP-based projectors~\citep{empat, background, fgamp} to connect different modules' input and output. Specifically, the PostProjector connects the Image Tokenizer output to the EmbeddingViT's input with a 2-layer MLP, whereas the VisProjector connects EmbeddingViT output to the LLM's input with a 3-layer MLP.

\paragraph{Training Task Proportion.} 
MMAR models $P(I|T)$, $P(T|I)$ and $P(I)$ through the conditional image generation task, the image understanding task and the unconditional image generation task, respectively. Through the first and the second stage training, we set the proportion of these three tasks as 50\%, 45\%, and 5\%, respectively. During the optional third training stage, only image generation tasks are applied, and the proportion of the conditional and unconditional image generation tasks are set to 90\% and 10\%, respectively.

\paragraph{Mixed-Precision Inference.} 
To enhance numerical stability during image generation, particularly when using small step sizes in the DDPM sampling process, we perform the model's forward pass in \texttt{bfloat16} (matching the training precision) but cast the output to \texttt{float32} before DDPM sampling. This mitigates potential rounding error without significant computational overhead, improving sampling accuracy efficiently.

\paragraph{Ablation-VQ.} Based on the MMAR-0.5B framework, we replace the Image Tokenizer from LDM-KL-16 to LDM-VQ-16. The image codes extracted using LDM-VQ-16 are then passed through a projector to increase the channel size to match the LLM's hidden size. Subsequently, we add a decoding Linear layer, which takes the hidden states of the LLM's output image portion as input and maps them to the image codebook. The Cross Entropy loss is then calculated between these mapped values and the ground-truth VQ codes.

\paragraph{Transfusion Reimplementation.} Our re-implementation version of Transfusion shares most of the MMAR-0.5B's model architecture except for the processing of the model input and output. Following the Transfusion paper, we adopt a input linear projection to convert the output of LDM-KL-16 image tokenizer into the LLM input representation, and use an output linear projection to convert the LLM output into the predicted noise. Noise of different levels are added to the image token input according to different pretraining tasks. For the image-to-text task, the diffusion time step is uniformly sampled within $t\in[0,500]$, while for the text-to-image task, the diffusion time step is uniformly sampled within $t\in[0,1000]$. A learnable time embedding corresponding to the time step $t$ is added after the input linear projection. MSE loss is calculated between the predicted and ground-truth noise. During inference, we treat the LLM as a diffusion model, with the condition being the concatenation of the text and the noisy image tokens of the previous diffusion time step $t+1$.

\subsection{Minimizing the Numerical Error in Diffusion Models}
\label{app:v-pred}

To make our discussion clearer, we switch the diffusion noise schedule into an angular form as follows:
\begin{equation}
  \begin{cases}
    \sin \phi_t = \sqrt{1-\bar{\alpha}_t}, \\
    \cos \phi_t = \sqrt{\bar{\alpha}_t}.
  \end{cases}
\end{equation}
In this way, the forward diffusion process can be written as follows:
\begin{equation}
  x^{(t)} = \sqrt{\bar{\alpha}_t}x + \sqrt{1-\bar{\alpha}_t}\epsilon = \cos \phi_t x + \sin \phi_t \epsilon, \label{eq:xt_angular}
\end{equation}
where $x^{(t)}$, $x$ and $\epsilon$ are noised image latent, original image latent and gaussian noise, respectively.

Our goal is to minimize the numerical error term in the DDIM sampling process. However, the form of DDIM sampling process is different under different parameterization method of the diffusion model. Therefore, we need to first define a general form to represent the diffusion model parameterization.

We consider the diffusion model output $u_\theta^{(t)}$ predict a linear combination of data $x$ and noise $\epsilon$, i.e. $u^{(t)} = a_t x + b_t\epsilon$. Note that the coefficients can vary according to the diffusion time step $t$.
Further re-writing the coefficients in the angular form gives:
\begin{equation}
  u^{(t)} = r_t\cos \psi_t x + r_t\sin \psi_t \epsilon, \label{eq:ut_angular}
\end{equation}
where $r_t = \sqrt{a_t^2 + b_t^2}$ represents the scale of $u^{(t)}$. $\cos \psi_t$ and $\sin \psi_t$ balance the proportion of $x$ and $\epsilon$.
Combining Eq.\ref{eq:xt_angular} and Eq.\ref{eq:ut_angular}, we can in turn represent $x$ and $\epsilon$ with $u^{(t)}$ and $x^{(t)}$:
\begin{equation}
  \label{eq:revert_x_eps}
  \begin{cases}
    x = \frac{\sin \psi_t x^{(t)} - \sin \phi_t u^{(t)}/r_t}{\cos \phi_t \sin \psi_t - \cos \psi_t \sin \phi_t} = \frac{\sin \psi_t x^{(t)} - \sin \phi_t u^{(t)}/r_t}{\sin(\psi_t - \phi_t)}, \\
    \epsilon = \frac{\cos \psi_t x^{(t)} - \cos \phi_t u^{(t)}/r_t}{\sin \phi_t \cos \psi_t - \sin \psi_t \cos \phi_t} = - \frac{\cos \psi_t x^{(t)} - \cos \phi_t u^{(t)}/r_t}{\sin(\psi_t - \phi_t)}.
  \end{cases}
\end{equation}
Now we consider the general form of DDIM sampling step~\citep{ddim}:
\begin{equation}
  x^{(t-1)} = \cos \phi_{t-1} \hat{x}_\theta(x^{(t)}) + \sin \phi_{t-1} \hat{\epsilon}_\theta(x^{(t)}),
\end{equation}
where $\hat{x}_\theta(x^{(t)})$ and $\hat{\epsilon}_\theta(x^{(t)})$ are the estimated image latent and noise, respectively. 

\begin{figure}
    \centering
    \includegraphics[width=0.8\linewidth]{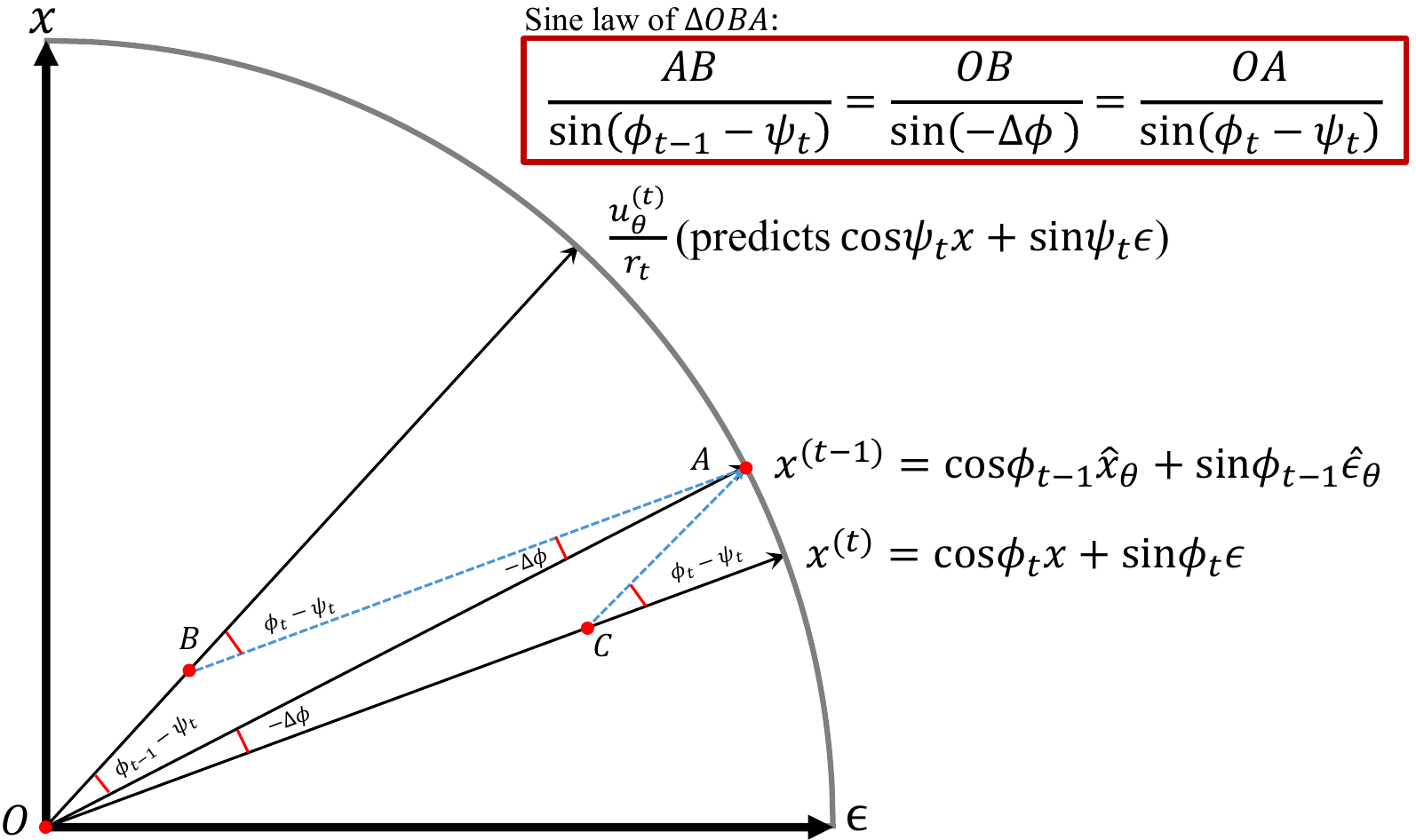}
    \caption{Geometric interpretation of a DDIM sampling step under arbitrary diffusion model parameterization.}
    \label{fig:geometric_understanding}
\end{figure}

Note that by using Eq.\ref{eq:revert_x_eps}, both of $\hat{x}_\theta(x^{(t)})$ and $\hat{\epsilon}_\theta(x^{(t)})$ can be derived from the noisy image latent $x^{(t)}$ and the diffusion model output $u_\theta^{(t)}$. Therefore, we can further represent $x^{(t-1)}$ in the following form:
\begin{align}
  x^{(t-1)} =& \cos \phi_{t-1} \frac{\sin \psi_t x^{(t)} - \sin \phi_t u_\theta^{(t)}/r_t}{\sin(\psi_t - \phi_t)} \notag\\
  &- \sin \phi_{t-1} \frac{\cos \psi_t x^{(t)} - \cos \phi_t u_\theta^{(t)}/r_t}{\sin(\psi_t - \phi_t)} \notag \\
  =& \frac{\sin (\phi_{t-1} - \phi_t) u_\theta^{(t)}/r_t - \sin (\phi_{t-1} - \psi_t) x^{(t)}}{\sin(\psi_t - \phi_t)}. \label{eq:general_ddim}
\end{align}
Eq.\ref{eq:general_ddim} represents the general form of DDIM sampling step under any kind of diffusion model parameterization in the form of Eq.\ref{eq:ut_angular}. To help understanding, we further present the geometric meaning of Eq.\ref{eq:general_ddim}. As shown in Fig.\ref{fig:geometric_understanding}, term $x^{(t-1)}, x^{(t)}$, and $u_\theta^{(t)}/r_t$ all locate on the unit circle in the $x-\epsilon$ plain. We find that Eq.\ref{eq:general_ddim} can be interpreted as projecting $x^{(t-1)}$ onto the $(x^{(t)},\frac{u_\theta^{(t)}}{r_t})$ coordinate system. We illustrate this projection by adding auxiliary line $AB$ and $AC$. By solving the sine law of $\triangle OBA$ given $OA=1$, we get:
\begin{equation}
  \begin{cases}
    OB = \frac{\sin(\Delta\phi)}{\sin(\psi_t - \phi_t)} \\
    BA = \frac{-\sin(\phi_{t-1}-\psi_t)}{\sin(\psi_t - \phi_t)}
  \end{cases}
\end{equation}
By representing $x^{(t-1)} = OB \cdot u_\theta^{(t)}/r_t + AB \cdot x^{(t)}$, we get:
\begin{equation}
  x^{(t-1)} = \frac{\sin(\Delta\phi)}{\sin(\psi_t - \phi_t)} u_\theta^{(t)}/r_t - \frac{\sin(\phi_{t-1}-\psi_t)}{\sin(\psi_t - \phi_t)}x^{(t)},
\end{equation}
which aligns with Eq.\ref{eq:general_ddim} given that $\Delta\phi = \phi_{t-1} - \phi_t$.

Now, we take the numerical error into consideration by multiplying the model output by a factor $1+\delta$, where $\delta$ represents the relative error:
\begin{equation}
  \Tilde{x}^{(t-1)} = \frac{\sin (\phi_{t-1} - \phi_t)(1+\delta)u_\theta^{(t)}/r_t - \sin (\phi_{t-1} - \psi_t) x^{(t)}}{\sin(\psi_t - \phi_t)}. \label{eq:general_ddim_with_numerr}
\end{equation}
Further, we can isolate the numerical error term from the ideal DDIM sampling step:
\begin{equation}
  \Tilde{x}^{(t-1)} = x^{(t-1)} + \sin (\phi_{t-1} - \phi_t) \frac{u_\theta^{(t)}/r_t}{\sin(\psi_t - \phi_t)}\delta. \label{eq:numerr_angular}
\end{equation}
From Eq.\ref{eq:numerr_angular}, we conclude that the numerical error of a DDIM sampling step is determined by four factors, namely, the step size $\Delta \phi = \phi_{t-1} - \phi_t$, the normalized model output $u_\theta^{(t)}/r_t$, the relative error of the data type $\delta$, and $\sin(\psi_t - \phi_t)$, which is decided by the parameterization of the diffusion model.

Notably, not all these four factors are useful to achieve the goal of minimizing the numerical error. For example, tuning down the step size only decreases the numerical error of each step. As a result, the total step number of DDIM sampling is increased proportionally, which cancels out the effect of error reduction of each single step. 
The factor $u_\theta^{(t)}/r_t$, which represents the normalized model prediction, is largely decided by the optimization progress and the target data distribution. An approximation can be derived from a perfect prediction, that is $u_\theta^{(t)}/r_t = u^{(t)}/r_t$. In this way,
\begin{align}
  \mathbb{E}[(u_\theta^{(t)}/r_t)^2] &\approx \mathbb{E}[(\cos \psi_t x + \sin \psi_t\epsilon)^2] \notag\\
  &=\cos^2 \psi_t\mathbb{E}[x^2]+\sin^2 \psi_t. 
\end{align}
In common practice, image tokens $x$ are normalized into unit standard deviation, leading to $\mathbb{E}[(u_\theta^{(t)}/r_t)^2]\approx\cos^2 \psi_t+\sin^2 \psi_t=1$. This is a constant number, meaning that there is not much potential to reduce the total numerical error via tuning down $u_\theta^{(t)}/r_t$.

If we decide to scale up our model, it is better to leverage the pre-trained LLMs as well as the highly efficient training infrastructure that is specifically optimized for LLMs. This makes \texttt{bfloat16} almost the only choice. As a result, the relative error $\delta$ is fixed to $1/128$.

Now, our only choice is to adjust the diffusion model parameterization method, so that $|\sin(\psi_t - \phi_t)|$ is maximized. A simple solution is to set $\psi_t - \phi_t = \pi / 2$, resulting in the following parameterization:
\begin{align}
  u^{(t)} &= r_t\cos (\phi_t + \pi / 2) x + r_t\sin (\phi_t + \pi / 2) \epsilon \notag\\
  &= r_t(\cos \phi_t \epsilon - \sin \phi_t x).\label{eq:optim_parmeterization_1}
\end{align}
Note that $r_t$ is still undetermined, which reflects the scale of $u^{(t)}$. From the analysis above, $r_t$ does not affect the numerical error term, since it is canceled out by the normalization of the model output, as seen in the factor $u_\theta^{(t)}/r_t$. Therefore, $r_t$ can be chosen freely, or based on other considerations. We consider that the smooth optimization of a neural network often requires the activation and output not too large or small. Therefore, we require a unit standard deviation for $u^{(t)}$, making $r_t=1$ constantly.

The final parameterization of our diffusion model is as follows:
\begin{equation}
  u^{(t)} = \cos \phi_t \epsilon - \sin \phi_t x.\label{eq:optim_parmeterization}
\end{equation}
We notice that this parameterization is coincidentally the ``v-prediction" parameterization~\citep{vpred}. Note that, however, ``v-prediction" is initially proposed for the efficient distillation of diffusion models, rather than reducing the numerical error of diffusion models. 
To the best of our knowledge, our work is the first to derive  ``v-prediction" parameterization from the first principle of minimizing the numerical error in diffusion models.

\begin{table*}[ht]
  \centering
  \vspace{-1em}
  \caption{Detailed visual understanding evaluation results.}
  \vspace{-1em}
  \fontsize{8pt}{8pt}\selectfont
  \begin{tabular}{l|ccc|ccccc}
    \toprule
    ~&~&~&~&~&\textbf{MMAR-0.5B}&\textbf{MMAR-0.5B}&~&\textbf{MMAR-7B}\\
     \multirow{-2}{*}{\textbf{Benchmark}}  & \multirow{-2}{*}{\textbf{Chameleon-7B}} & \multirow{-2}{*}{\textbf{Transfusion*}} & \multirow{-2}{*}{\textbf{Show-o}} & \multirow{-2}{*}{\textbf{MMAR-0.5B}} &  \textbf{w/ $\epsilon$-pred.}  &  \textbf{w/ VQ} & \multirow{-2}{*}{\textbf{MMAR-7B}} & \textbf{w/ Stg3}\\ 
    \midrule
    \rowcolor{gray!10}
      \textbf{AI2D}~\citep{ai2d}           & 34.81 & 40.22 & 32.48 & 43.43 & 41.90 & 41.90 & \textbf{64.64} & 63.54 \\
      \textbf{ChartQA}~\citep{chartqa}	    & 3.84  & 9.56  & 11.32 & 10.20 & 10.36 & 9.36  & \textbf{13.64} & 12.52 \\
    \rowcolor{gray!10}
      \textbf{DocVQA}~\citep{docvqa}	        & 1.51  & 6.72  & \textbf{18.24} & 7.62  & 6.77  & 6.79  & 11.12 & 10.62 \\
      \textbf{Hallu.Bench}~\citep{hallubench}    & 39.01 & 41.54 & 40.90 & 42.80 & 41.11 & 41.54 & \textbf{53.10} & 52.05 \\
    \rowcolor{gray!10}
      \textbf{MathVista}~\citep{mathvista}      & 21.90 & 22.60 & 23.20 & 21.60 & 23.10 & 22.90 & \textbf{32.40} & 31.30 \\
      \textbf{MMBench}$^{CN}$~\citep{mmbench} & 10.14 & 27.23 & 0.52  & 43.99 & 38.83 & 31.87 & \textbf{70.53} & 66.75 \\
    \rowcolor{gray!10}
      \textbf{MMBench}$^{EN}$~\citep{mmbench} & 13.32 & 29.47 & 42.44 & 48.45 & 45.53 & 37.54 & \textbf{70.45} & 67.78 \\
      \textbf{MME}$^P$~\citep{mme}        & 125.8 & 594.3 & 1182.7& 882.1 & 880.7 & 618.2 & \textbf{1486.9} & 1421.9  \\
    \rowcolor{gray!10}
      \textbf{MME}$^C$~\citep{mme}        & 33.9 	& 206.1 & 225.0	& 256.8 & 232.1 & 273.2 & 268.9 & \textbf{303.6}   \\
      \textbf{MMMU}~\citep{mmmu}           & 24.00 & 29.33 & 26.44 & 29.33 & 25.33 & 29.67 & 41.33 & \textbf{47.33} \\
    \rowcolor{gray!10}
      \textbf{MMStar}~\citep{mmstar}         & 20.47 & 28.13 & 32.00 & 32.13 & 31.07 & 28.07 & \textbf{41.87} & 40.87 \\
      \textbf{MMVet}~\citep{mmvet}          & 7.34  & 13.90 & 20.87 & 18.49 & 17.98 & 14.45 & \textbf{30.64} & 29.17 \\
    \rowcolor{gray!10}
      \textbf{OCRBench}~\citep{ocrbench}       & 0.50  & 2.30  & 15.20 & 18.70 & 7.10  & 2.10  & \textbf{25.80} & 21.20 \\
      \textbf{POPE}~\citep{pope}           & 30.86 & 66.90 & \textbf{84.50} & 70.74 & 71.14 & 66.98 & 84.02 & 83.21 \\
    \rowcolor{gray!10}
      \textbf{RealWorldQA}    & 27.06 & 36.99 & 27.97 & 38.30 & 35.16 & 36.60 & \textbf{53.59} & 52.68 \\
      \textbf{ScienceQA}~\citep{scienceqa}      & 44.83 & 45.92 & 41.82 & 47.54 & 45.21 & 45.35 & \textbf{74.39} & 73.20 \\
    \rowcolor{gray!10}
      \textbf{SEEDBench}~\citep{seedbench}      & 34.61 & 42.40 & 51.61 & 55.70 & 53.72 & 44.93 & \textbf{68.63} & 66.59 \\
      \textbf{TextVQA}~\citep{textvqa}        & 5.43  & 9.94  & \textbf{38.35} & 16.77 & 12.40 & 9.46  & 24.37 & 21.35 \\
    \midrule
    \rowcolor{gray!10}
      \textbf{Average}        & 18.34 & 28.26 & 33.06 & 34.56 & 32.21 & 29.70 & \textbf{48.25} & 47.18 \\
    \bottomrule
  \end{tabular}
  \vspace{-1em}
  \label{tab:detailed_visual_und}
\end{table*}

\subsection{Deriving Theoretical Numerical Error for $\epsilon$-Prediction Models}
\label{app:theoretical_error}
The $\epsilon$-prediction parameterization corresponds to $\psi_t=\frac{\pi}{2}$ in the angular parameterization form given by Eq.\ref{eq:ut_angular}.
Substituting $\psi_t=\frac{\pi}{2}$ and $u^{(t)}_\theta/r_t=\epsilon_\theta$ into Eq.\ref{eq:numerr_angular}, we get:
\begin{equation}
  \Tilde{x}^{(t-1)} = x^{(t-1)} + \sin (\phi_{t-1} - \phi_t) \frac{\epsilon_\theta}{\cos(\phi_t)}\delta. \label{eq:numerr_angular_npred}
\end{equation}
Further, we cancel out the step size factor $\sin (\phi_{t-1} - \phi_t)$ within the numerical error term, only focusing on ``the numerical error introduced per \textbf{unit} DDIM step":
\begin{equation}
  e^{(t)} = \frac{\epsilon_\theta}{\cos \phi_t}\delta. \label{eq:unit_error}
\end{equation}
Next, we will show that $e^{(t)}$ can also be interpreted as the equivalent v-prediction numerical error for an $\epsilon$-prediction model.

For an $\epsilon$-prediction model, $u_\theta^{(t)} = \epsilon_\theta$. In order to calculate the equivalent $v^{(t)}_\theta$ value, we need to represent $v^{(t)}_\theta$ with the predicted $\epsilon_\theta$ and the known $x^{(t)}$, which is calculated as follows:
\begin{align}
  v^{(t)}_\theta &= \cos \phi_t \epsilon_\theta - \sin \phi_t \hat{x}_\theta(x^{(t)}) \notag\\
  &= \cos \phi_t \epsilon_\theta - \sin \phi_t \frac{x^{(t)} - \sin \phi_t \epsilon_\theta}{\cos \phi_t} \notag\\
  &= \frac{\epsilon_\theta}{\cos \phi_t} - \tan \phi_t x^{(t)}. 
\end{align}
Considering the numerical error, we get:
\begin{equation}
  \Tilde{v}^{(t)}_\theta = \frac{\epsilon_\theta(1+\delta)}{\cos \phi_t} - \tan \phi_t x^{(t)} = v^{(t)}_\theta + \frac{\epsilon_\theta}{\cos \phi_t}\delta. 
\end{equation}
Note that the numerical error term in the above equation is exactly $e^{(t)}$, proving that $e^{(t)}$ can be interpreted as the equivalent v-prediction numerical error for an $\epsilon$-prediction model.

Taking numerical error effect into the v-prediction-based diffusion loss, we get:
\begin{align}
  \mathbb{E}[(v^{(t)} - \Tilde{v}^{(t)}_\theta)^2] =& \mathbb{E}[(v^{(t)} - v^{(t)}_\theta - e^{(t)})^2] \notag\\
  =& \mathbb{E}[(v^{(t)} - v^{(t)}_\theta)^2] - 2\mathbb{E}[(v^{(t)} - v^{(t)}_\theta)e^{(t)}] \notag\\
  &+ \mathbb{E}[(e^{(t)})^2]. \label{eq:theoretical_numerical_error_loss_overhead}
\end{align}
Due to the fact that numerical error $e^{(t)}$ is independent from the training loss and that the expectation of $e^{(t)}$ is 0, we get $\mathbb{E}[(v^{(t)} - v^{(t)}_\theta)e^{(t)}]=0$.
Therefore, the only numerical error term is $\mathbb{E}[(e^{(t)})^2]$. Given that the standard deviation of $\epsilon_\theta$ is 1, and considering that we use \texttt{bfloat16} as training data type, which means $\delta=1/128$, we get 
\begin{equation}
    \mathbb{E}[(e^{(t)})^2] = 1/(128\cos(\phi_t))^2 = 1/(128^2\bar{\alpha_t}).
\end{equation}
This is the theoretical numerical error of the v-prediction diffusion loss for an $\epsilon$-prediction model.

\begin{figure}[H]
    \centering
    \vspace{-0.7em}
    \includegraphics[width=0.70\linewidth]{imgs/cfg.pdf}
    \caption{The impact of CFG scale on image generation quality.}
    \label{fig:cfg_analyze}
    \vspace{-1em}
\end{figure}

\subsection{CFG With $v$-prediction}\label{appendix:CFG_V}
From Equation $ v^{(t)}_i = \sqrt{\bar{\alpha}_t}\mathbf{\epsilon} - \sqrt{1 - \bar{\alpha}_t}x_i$, we can derive the following equation.
\begin{align}
     \epsilon = \sqrt{1-\bar{\alpha}^{(t)}}x^{(t)} + \sqrt{\bar{\alpha}^{(t)}}v
\end{align}
For the CFG of $\epsilon$, it can be simplified as follows.
\begin{align*}
     \epsilon = & \epsilon_u + \omega(\epsilon_c-\epsilon_u) \\
       = & \sqrt{1-\bar{\alpha}^{(t)}}x^{(t)} + \sqrt{\bar{\alpha}^{(t)}}v_u + \omega\sqrt{\bar{\alpha}^{(t)}}(v_c-v_u) \\
     =& \sqrt{1-\bar{\alpha}^{(t)}}x^{(t)} + \sqrt{\bar{\alpha}^{(t)}}(v_u + \omega(v_c-v_u)) \tag{\theequation}\stepcounter{equation}
\end{align*}
Ultimately, we obtain $v=v_u+\omega(v_c-v_u)$. The CFG of $v$ and $\epsilon$ are equivalent.

\begin{figure*}[!h]
    \centering
    \includegraphics[width=0.9\linewidth]{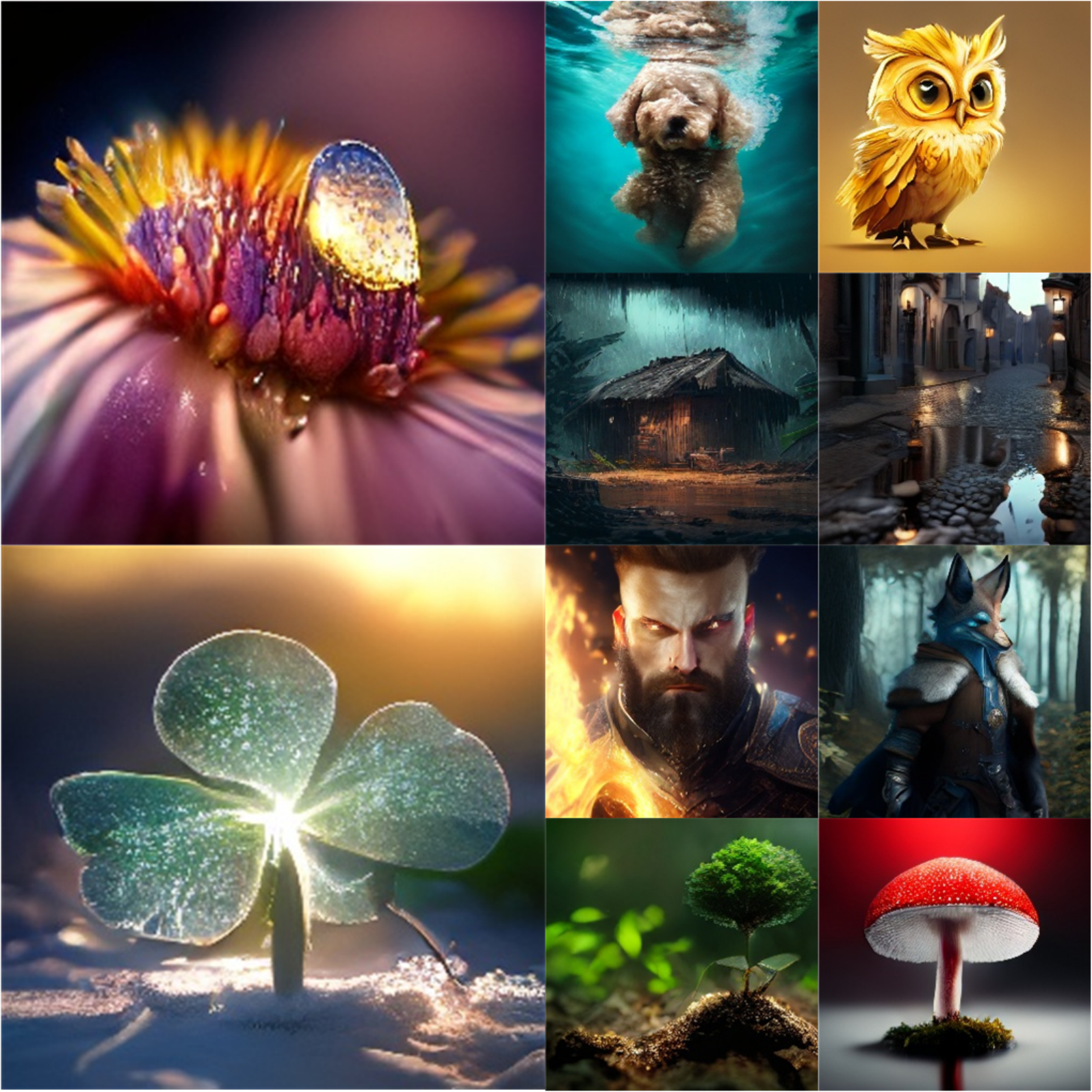}
    \caption{Generated images from MMAR-7B after the third training stage (1/2).}
    \label{fig:gen_sample}
    \vspace{-1em}
\end{figure*}

\subsection{Impact of CFG Scaling}
We select models from the second and fourth epochs of the first stage as starting points for the second stage, train them for 3 epochs, and then test the MSCOCO FID-30K under varying CFG intensities. As shown in Fig. \ref{fig:cfg_analyze}, our method achieves better FID scores as the CFG scale increases from 1 to 10. It is worth noting that most probabilistic generative models typically have a CFG scale between 1.5 and 5. 
Additionally, it is observed that a longer training duration in the first stage consistently results in better generation outcomes at all CFG scales.

\subsection{Detailed Visual Understanding Evaluation Results}
\label{app:details_understand}
A total of 18 visual understanding benchmarks from VLMEvalKit~\citep{vlmevalkit} are used to evaluate MMAR models comprehensively. The evaluation is also conducted on the existing joint image-text probabilistic models using the publicly available checkpoints\footnote{https://huggingface.co/facebook/chameleon-7b}\footnote{https://huggingface.co/showlab/show-o-w-clip-vit}. The detailed evaluation results are shown in Table \ref{tab:detailed_visual_und}. All scores have been scaled to a range of 0 to 100 except that we show the original score of MME benchmarks. The average score is calculated on the normalized score of all the benchmarks including MME.

\subsection{Examples:Image Generation Sample} \label{app:gen_sample}
\begin{figure*}[!h]
    \centering
    \includegraphics[width=0.9\linewidth]{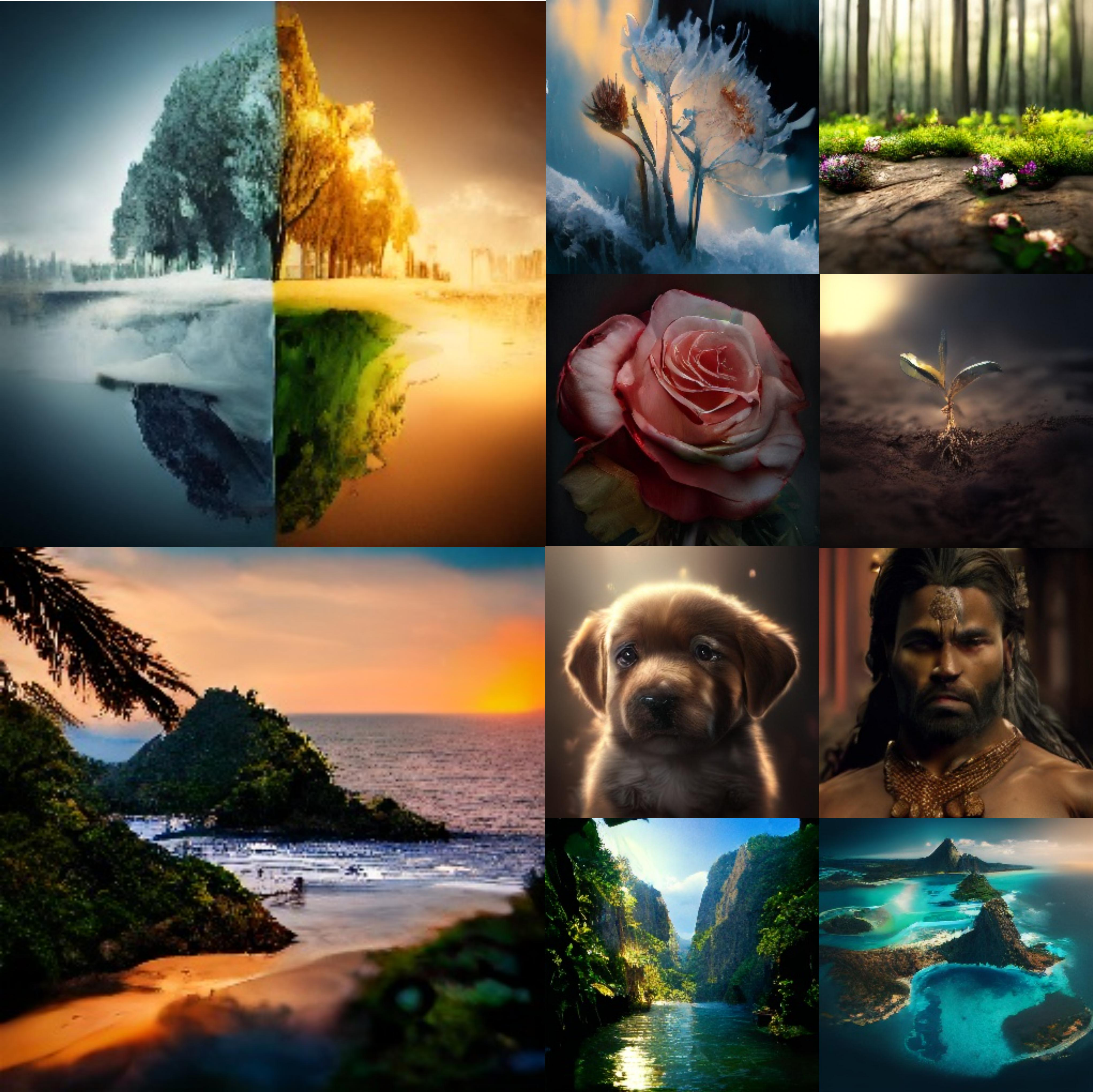}
    \caption{Generated images from MMAR-7B after the third training stage (2/2).}
    \label{fig:gen_sample1}
    \vspace{-1em}
\end{figure*}

In Fig.\ref{fig:gen_sample},~\ref{fig:gen_sample1}, we showcase some generated examples from MMAR-7B after the third stage training, featuring animals, plants, real-world scenes, artistic scenes, and counterfactual themes.

\subsection{Numerical Error Empirical Validation}\label{app:demo_numerical}
\label{app:numerical_Error}

\begin{figure}[ht]
    \centering
    \includegraphics[width=0.9\linewidth]{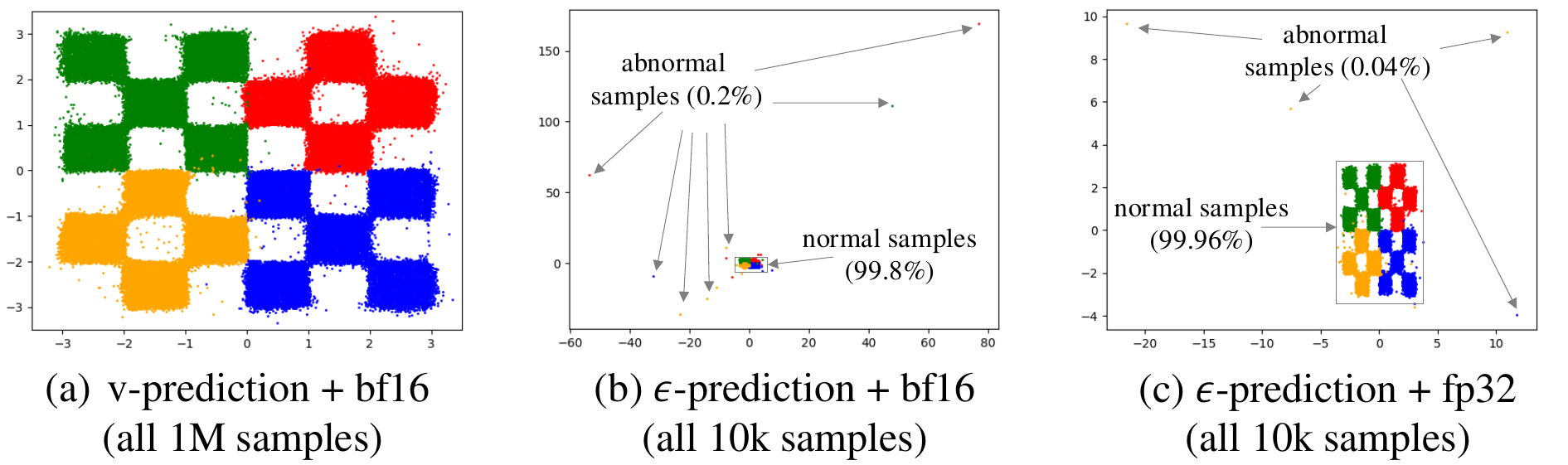}
    \caption{
    Diffusion sampling results with different prediction targets and training precision. Figures are scaled differently to include all generated samples.
    }
    \label{fig:toy_example}
\end{figure}

We provide direct empirical demonstration using a 2D chessboard distribution (class-conditioned quadrant mapping). As Fig.\ref{fig:toy_example} reveals: (1) v-prediction maintains stability even with bf16 precision at scale (1M samples, Fig.(a)), achieving $< 10^{-6}$ token error rate. (2) $\epsilon$-prediction exhibits visible artifacts (Fig.(b),(c)) with 0.2\% token error rate (bf16), equivalent to 51.2\% image-level failure for 256-token images. (3) Precision elevation (fp32) reduces artifacts to 0.04\%. These observations validate v-prediction's critical role in mitigating low-precision training risks.